\newif\ifdraft
\draftfalse

\documentclass{iopjournal}
\usepackage{natbib}

\usepackage{amssymb}
\usepackage{amsmath}

\usepackage{algorithm,algorithmic}
\usepackage{amsfonts}
\usepackage{hyperref}
\usepackage{bbm}
\usepackage{graphicx}
\usepackage{stfloats}  
\usepackage{caption,subcaption}
\usepackage{multirow}

\usepackage{comment}
\usepackage{hhline}
\usepackage{todonotes}
\usepackage{siunitx}
\usepackage{booktabs}
\usepackage{csvsimple}
\usepackage{datatool}
\usepackage{array}
\usepackage{xcolor}

\usepackage{textcomp}
\usepackage{makecell}
\usepackage{helvet}
\usepackage{hyperref}
\hypersetup{hidelinks=true}
\usepackage{cleveref}

\newcommand{\heading}[1]{\noindent\textbf{#1}}

\newcommand\colnst[1]{{\color{red}#1}}
\newcommand\coltw[1]{{\color{blue}#1}}

\usepackage[normalem]{ulem} 
\newcommand{\stkout}[1]{\ifmmode\text{\sout{\ensuremath{#1}}}\else\sout{#1}\fi}
\newcommand{\citepx}[1]{\mbox{\citep{#1}}}
\newcommand{\citeppx}[1]{\mbox{\citepp{#1}}}
\newcommand{\citeptx}[1]{\mbox{\citept{#1}}}
\ifdraft
\newcommand{\added}[1]{\textcolor{blue}{#1}}
\newcommand{\deleted}[1]{\textcolor{red}{\stkout{#1}}}
\newcommand{\replaced}[2]{\textcolor{blue}{#1} \textcolor{red}{\stkout{#2}}}
\newcommand{\deletedfloat}[1]{}
\newcommand{\commented}[1]{\textcolor{blue}{#1}}
\else
\newcommand{\added}[1]{#1}
\newcommand{\deleted}[1]{}
\newcommand{\replaced}[2]{#1}
\newcommand{\deletedfloat}[1]{}
\newcommand{\commented}[1]{}
\fi

\begin{document}

\articletype{Paper}


\title{Multi-Window Temporal Analysis for Enhanced Arrhythmia Classification: Leveraging Long-Range Dependencies in Electrocardiogram Signals}

\author{Tiezhi Wang$^1$, Wilhelm Haverkamp$^2$ and Nils Strodthoff$^{1,*}$}

\affil{$^1$AI4Health Department, Oldenburg University, Oldenburg, Germany}

\affil{$^2$Department of Cardiology, Angiology and Intensive Care Medicine, Charité Campus Mitte, German Heart Center of the Charité-University Medicine Berlin, Berlin, Germany}

\affil{$^*$Author to whom any correspondence should be addressed.}

\email{nils.strodthoff@uol.de}

\begin{abstract}
\added{\textit{Objective.}} Arrhythmia classification from electrocardiograms (ECGs) suffers from high false positive rates and limited cross-dataset generalization, particularly for atrial fibrillation (AF) detection where specificity ranges from 0.72 to 0.98 using conventional 30-second analysis windows. While conventional deep learning approaches analyze isolated 30-second ECG windows, many arrhythmias, particularly atrial fibrillation and atrial flutter, exhibit diagnostic features that emerge over extended time scales. \added{\textit{Approach.}} We introduce S4ECG, a deep learning architecture based on structured state-space models (S4), designed to capture long-range temporal dependencies by jointly analyzing multiple consecutive ECG windows spanning up to 20 minutes.
We evaluated S4ECG on four publicly available databases for multi-class arrhythmia classification, including systematic cross-dataset evaluations to assess out-of-distribution robustness. \added{\textit{Main results.}} Multi-window analysis consistently outperformed single-window approaches across all datasets, improving the macro-averaged area under the receiver operating characteristic curve (AUROC) by 1.0-11.6 percentage points. For AF detection specifically, specificity increased from 0.718-0.979 (single-window) to 0.967-0.998 (multi-window) at a fixed sensitivity threshold, representing a 3-10 fold reduction in false positive rates. \added{\textit{Significance.}} Comparative analysis against convolutional neural network baselines demonstrated superior performance of the S4 architecture. Cross-dataset evaluation revealed that multi-window approaches substantially improved generalization performance, with smaller performance degradation when models were tested on held-out datasets from different institutions and acquisition protocols. A systematic investigation revealed optimal diagnostic windows of 10-20 minutes, beyond which performance plateaus or degrades. These findings demonstrate that structured incorporation of extended temporal context enhances both arrhythmia classification accuracy and cross-dataset robustness. The identified optimal temporal windows provide practical guidance for ECG monitoring system design and may reflect underlying physiological timescales of arrhythmogenic dynamics.

\end{abstract}




\section{Introduction}
\label{sec:introduction}

\heading{Clinical burden of cardiac arrhythmias}
Cardiovascular diseases remain the leading cause of mortality worldwide, with arrhythmias representing a significant subset of these conditions that can lead to sudden cardiac death, stroke, and heart failure if left undetected and untreated~\citep{ansari2023deep, svennberg2025state, ko2025atrial}. The clinical landscape is experiencing a notable shift toward atrial fibrillation (AF) as the most prevalent sustained arrhythmia, affecting millions of patients globally and imposing substantial healthcare burdens~\citep{conrad2025changing}. Early and accurate detection of arrhythmias is crucial for timely intervention~\citep{ansari2023deep, ko2025atrial}, with continuous electrocardiographic monitoring playing an increasingly vital role in modern cardiology practice~\citep{svennberg2025state}.

\heading{Challenges in AF detection} The advent of portable devices and remote monitoring technologies has revolutionized arrhythmia detection~\citep{ko2025atrial, abdelrazik2025wearable}, enabling long-term continuous monitoring outside traditional clinical settings. However, the vast amounts of data generated by these devices present significant challenges for manual interpretation, creating an urgent need for automated algorithms that can reliably and accurately identify arrhythmic episodes. A particular challenge for arrhythmia detection algorithms is the high rate of false positive alarms, accounting, for example, for almost 60\% of the overall remote transmissions from implantable loop recorders~\citep{covino2024false}. While substantial progress has been made in automated ECG analysis research, enhancing model performance remains a pressing issue, particularly given the temporal complexity and variability inherent in cardiac rhythm disturbances that unfold over extended time periods.

\heading{Long-range correlations}
The cardiovascular system exhibits well-documented long-range temporal correlations, particularly evident in heart rate variability patterns during different physiological states~\citep{bunde2000correlated}. These long-range interactions manifest across multiple timescales, from beat-to-beat variations to circadian rhythms, and have been shown to carry diagnostic information for cardiac pathology detection~\citep{agliari2020detecting}. For instance, healthy heart dynamics exhibit multifractal complexity persisting for at least 700 beats (approximately 10 minutes) - a hallmark of physiological control markedly reduced in cardiac pathology~\citep{ivanov1999multifractality}. Similarly, detrended fluctuation analysis of 24-hour heartbeat recordings reveals that the characteristic scale-invariant long-range correlations in healthy subjects persist across scales from $10^2$ to $10^3$ beats (approximately 1 to 20 minutes), whereas pathologic dynamics deviate from this behavior at these scales~\citep{peng1995quantification}. Additionally, cardiac electrophysiology and autonomic tone exhibit pronounced circadian rhythms, and arrhythmic events cluster by time of day~\citep{kelters2025circadian}. The presence of such temporal dependencies suggests that arrhythmia detection algorithms could benefit substantially from incorporating extended temporal context.

\heading{Algorithmic approaches}
Automated ECG analysis has been an active area of research for several decades, with traditional approaches focusing primarily on handcrafted feature extraction and classical machine learning algorithms. Early methods relied on morphological features, frequency domain characteristics, and statistical measures derived from beat-to-beat intervals~\citep{Murat2021}. These approaches, while providing interpretable results, often struggled with the variability inherent in real-world ECG recordings and required extensive domain expertise for feature engineering. The development of deep learning has transformed ECG analysis, with convolutional neural networks (CNNs) and (to a lesser extent) long short-term memory (LSTM) networks, emerging as the dominant paradigm for automated arrhythmia detection~\citep{Murat2021}.

\heading{Shortcomings of existing approaches}
However, most existing approaches have primarily focused on single-window classification, where individual 5-30 second \emph{windows} are analyzed independently to identify the underlying cardiac rhythm~\citep{Murat2021}. This paradigm, while computationally efficient and conceptually straightforward, inherently limits the temporal context available for decision-making. Single-window models cannot capture rhythm transitions, paroxysmal episodes, or gradual changes in cardiac rhythm that may unfold over minutes to hours~\citep{ko2025atrial}—temporal patterns that are clinically significant for accurate arrhythmia characterization. Single-window models tend to misjudge contiguous arrhythmic events by making inconsistent predictions across adjacent segments, leading to inappropriate rhythm segmentation boundaries and fragmented episode detection. This segmentation error stems from the lack of temporal overview that would enable recognition of sustained arrhythmic patterns extending beyond individual windows. Furthermore, most existing ECG analysis architectures are fundamentally constrained by their reliance on traditional deep learning components, primarily CNNs and LSTM~\citep{Murat2021}. While CNNs excel at capturing local morphological features within individual heartbeats, they are inherently limited in modeling long-range temporal dependencies that extend beyond their receptive field. LSTMs, despite their theoretical ability to capture sequential dependencies, suffer from vanishing gradient problems and computational inefficiency when processing extended sequences, making them impractical for multi-window analysis spanning tens of minutes. These architectural limitations have prevented the field from leveraging recent advances in sequence modeling, particularly structured state space models (S4)~\citep{Gu2021EfficientlyML} , which offer superior capabilities for efficient long-range dependency capture.

\heading{Contributions}
This work introduces S4ECG, a hierarchical deep learning architecture that systematically explores the impact of long-range temporal interactions for arrhythmia detection.

Our primary contribution lies in adapting and extending the encoder-predictor paradigm from sleep staging~\citep{wang2025s4sleep} to ECG analysis, employing S4 models at both window-level and sequence-level processing stages to enable efficient capture of long-range dependencies. Firstly, we confirm the superiority of S4-based encoders over widely used CNN-based encoders in line with prior work \citep{al2025benchmarking}. Secondly and most importantly, we demonstrate that multi-window models consistently outperform single-window approaches across diverse datasets and evaluation scenarios.

We present the first comprehensive study to systematically investigate the optimal temporal window for ECG arrhythmia detection, evaluating model performance across temporal contexts ranging from 2 to 60 windows, i.e., 1 to 30 minutes of continuous ECG data at a window length of 30 seconds. \added{We emphasize that we work with a fixed window size of 30s throughout this work, longer temporal context is implemented by considering multiple consecutive windows at once.} Our findings reveal consistent optimal performance in the 20-40 window range, i.e., 10 to 20 minutes, suggesting fundamental characteristics of cardiac rhythm analysis that extend beyond dataset-specific artifacts.
Our evaluation encompasses training on large-scale and medium-sized datasets, followed by rigorous out-of-distribution testing on medium-sized and smaller benchmark databases from PhysioNet~\citep{goldberger2000physiobank}. This comprehensive evaluation framework spans diverse acquisition protocols, patient populations, and clinical contexts, enabling thorough assessment of model robustness and generalizability across real-world deployment scenarios. The consistency of our findings across these diverse datasets provides strong evidence for the generalizability of multi-window approaches in clinical ECG analysis.
To summarize, we put forward the following technical contributions:
\begin{enumerate}
    \item We assess hierarchical prediction models leveraging structured state space models that showed outstanding performance in sleep stage prediction~\citep{wang2025s4sleep} for the purpose of arrhythmia prediction. The proposed models, in particular when trained on large-scale datasets such as Icentia11k~\citep{icentia11kphysionet}, show strong predictive performance, also when evaluated on out-of-distribution data.
    \item We provide robust evidence for the advantages of jointly predicting multiple prediction windows at once as opposed to a single window at a time, as predominantly considered in the literature, both in terms of in-distribution and out-of-distribution performance. We present qualitative evidence for the advantages of such multi-window prediction models.
\end{enumerate}

\section{Methods}
\label{sec:methods}

\subsection{Datasets and experimental setup}

\heading{Datasets} This study investigates the effectiveness of multi-window deep learning models for arrhythmia detection using a comprehensive evaluation framework encompassing training, in-distribution (ID), and out-of-distribution (OOD) assessments. We use two datasets for model training and ID evaluation, and—for OOD evaluation—combine one of these training datasets with two additional external datasets to assess robustness and generalizability. All datasets are publicly accessible through PhysioNet~\citep{goldberger2000physiobank}, ensuring reproducibility.
Table~\ref{tab:dataset_summary} and \ref{tab:rhythm_distribution} provide an overview of the four datasets, including patient counts, recording durations, sampling frequencies, and rhythm-type distributions. The substantial differences in dataset scale, temporal resolution, and class prevalence enable a thorough evaluation of multi-window performance under varying conditions, from ID settings that mirror training characteristics to challenging OOD scenarios. \added{We emphasize that all deep learning models were trained directly on the raw ECG time-series signals, without applying additional preprocessing steps such as baseline-wander correction or signal filtering. This design choice was intentional: it enables the model to learn directly from raw signals, thereby improving generalizability across different acquisition systems and avoiding potential information loss from aggressive filtering or hand-crafted preprocessing pipelines.} Detailed dataset and preprocessing descriptions are provided in Section~\ref{app:datasets}.

\begin{table}[!h]
    \centering
    \caption{Summary of ECG datasets used in this study.}
    \label{tab:dataset_summary}
    \resizebox{\columnwidth}{!}{
    \begin{tabular}{cccccc}
        \toprule
        \textbf{Dataset} & \textbf{Use} & \textbf{Patients} & \textbf{Sampling rate (Hz)} & \textbf{Duration per recording} & \textbf{Total hours} \\
        \midrule
        Icentia11k\citep{icentia11kphysionet} & Training, ID eval & 11,000 & 250 & Variable (hours-weeks) & $\sim$110,000 \\
        \midrule
        LTAFDB\citep{ltafdb} & Training, ID/OOD eval & 84 & 128 & 24-25 hours & $\sim$2,000 \\
        \midrule
        AFDB\citep{afdb} & OOD eval & 25 & 250 & $\sim$10 hours & $\sim$250 \\
        \midrule
        MITDB\citep{mitdb} & OOD eval & 47 & 360 & 30 minutes & $\sim$24 \\
        \bottomrule
    \end{tabular}}
\end{table}

\heading{Experimental setup} We convert rhythm annotation timestamps into relative fractions of the considered rhythm types per prediction window, see Section~\ref{app:datasets}. We use these fractional labels as prediction targets using a binary crossentropy loss function optimized through an AdamW optimizer. The macro-averaged area under the receiver operating characteristic curve (AUROC) serves as primary metric. For clinical interpretability, we also report AF detection specificity at a fixed sensitivity of 0.9, consistent with the performance levels achieved by FDA-cleared wearable AF detection devices~\citep{perez2019large,bumgarner2018smartwatch}. We refer the reader to Section~\ref{app:setup} for extensive details on the experimental setup.

\begin{table}[!h]
    \centering
    \caption{Rhythm-type distribution across datasets (\% of windows).}
    \label{tab:rhythm_distribution}
    \resizebox{\columnwidth}{!}{
    \begin{tabular}{ccccc}
        \toprule
        \textbf{Rhythm type} & \textbf{Icentia11k} & \textbf{LTAFDB} & \textbf{AFDB} & \textbf{MITDB} \\
        \midrule
        Normal (N) & 87.2\% & 42.1\% & 56.8\% & 78.3\% \\
        Atrial fibrillation (AF) & 11.4\% & 51.2\% & 38.7\% & 19.2\% \\
        Atrial flutter (AFLT) & 1.4\% & -- & 4.5\% & 2.5\% \\
        Supraventricular tachyarrhythmia (SVTA) & -- & 6.7\% & -- & -- \\
        \bottomrule
    \end{tabular}}
\end{table}

\subsection{S4ECG architecture}
This work builds on prior advances in sleep staging~\citep{wang2025s4sleep}, which conducted extensive architecture searches for long time-series classification. It provided evidence for the superiority of S4 layers over LSTMs or transformer architectures in a closely related setting. We therefore refrain from excessive ablation studies and focus exclusively on the best-performing model identified in prior work. The S4ECG model implements an encoder-predictor paradigm that leverages S4 layers as core components across two processing stages.

\heading{Architecture overview} The model processes multi-window ECG sequences through a hierarchical encoder-predictor design.
An input sequence of length $L$ samples is first segmented into $N$ non-overlapping windows of fixed length $L_{\text{window}} = 3840$ samples (30 seconds at 128~Hz), where $N = L / L_{\text{window}}$. Each window is encoded independently; the resulting token sequence is then modeled by a predictor to capture inter-window dependencies. A classification head produces per-window rhythm predictions. \added{Detailed architectural and training hyperparameters are provided in Section~\ref{app:repro}.}

\heading{Window-level encoder} Each 30-second window is passed through a convolutional front-end (two 1D convolutional layers with 128 channels, kernel size 3, and stride 2), which reduces the temporal dimension from 3840 to 960 samples, followed by an S4 stack (model dimension 512, state dimension 64, 4 layers, bidirectional). A pooling operation compresses each window to a single 512-dimensional token.

\heading{Multi-window predictor} The sequence of window tokens is processed by a second, four-layer S4 module (model dimension 512, bidirectional) that captures long-range temporal dependencies across windows. The output is fed to a linear classification head to produce rhythm predictions.

\begin{figure*}[!htbp]
    \centering
    \begin{subfigure}[b]{0.475\columnwidth}
        \centering
        \includegraphics[width=\columnwidth]{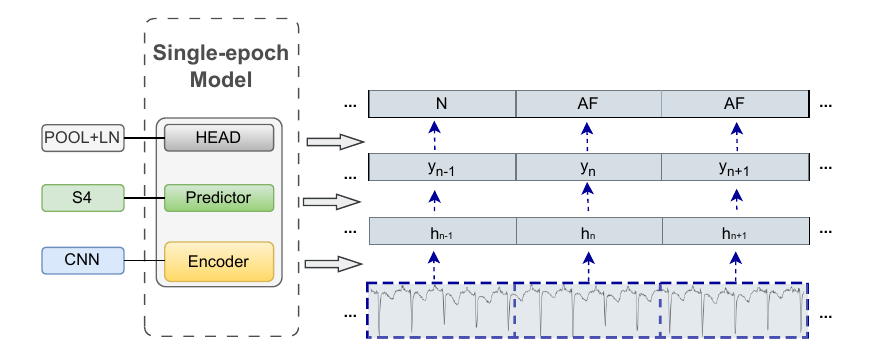}
        \caption{Single-window architecture for baseline comparison.}
        \label{fig:arc_single}
    \end{subfigure}
    \begin{subfigure}[b]{0.475\columnwidth}
        \centering
        \includegraphics[width=\columnwidth]{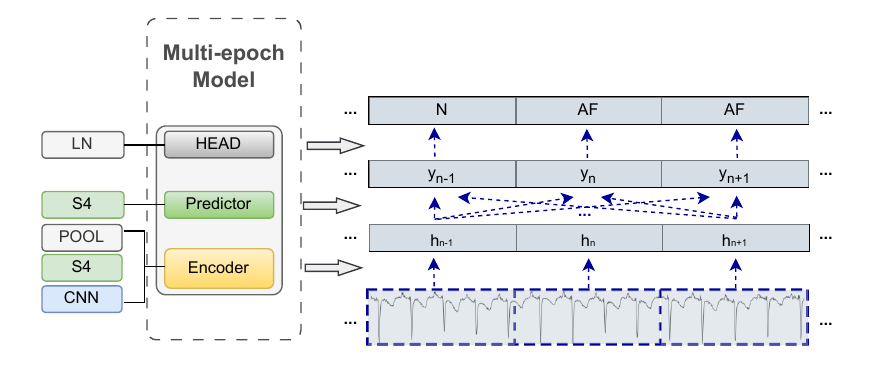}
        \caption{Multi-window S4ECG architecture with hierarchical encoder-predictor design}
        \label{fig:arc_multi}
    \end{subfigure}
    
    \caption{Architecture comparison: (a) single-window baseline model and (b) multi-window S4ECG model with temporal dependency modeling.}
    \label{fig:architectures}
\end{figure*}

\heading{Multi-window vs. single-window comparison} Unlike conventional single-window models that process fixed 30-second segments (input size = 3,840 samples), as shown in Figure~\ref{fig:arc_single}, our S4ECG model shown in Figure~\ref{fig:arc_multi} processes variable-length sequences containing multiple windows. For example, with input size 38,400, the model processes $N = 10$ windows spanning 5 minutes of continuous ECG. This enables modeling of patterns extending beyond individual segments, such as paroxysmal episodes and rhythm transitions.

The inclusion of the multi-window predictor and the formulation of the task as a joint prediction over several windows constitute the central novelty of our approach. We systematically compare this against conventional single-window models across temporal contexts. For LTAFDB, we evaluate 2 to 60 windows (1 to 30 minutes of ECG); for the large-scale Icentia11k dataset, we narrow this to 10 to 60 windows (5 to 30 minutes) for computational efficiency. 

\heading{Single-window baselines} We consider two baseline models that operate on a single window as input. On the one hand, we consider a xResNet1d50 model \citep{strodthoff2020deep} as representative for the predominantly used CNNs for this task. On the other hand, we consider a S4-based single-window baseline model, which emerged as strongest single-window backbone in \citep{wang2025s4sleep}.

\section{Results}
\label{sec:results}

We present a comprehensive evaluation of the S4ECG multi-window model across both ID and OOD scenarios. Our results demonstrate consistent and substantial improvements of multi-window models over conventional single-window approaches across all evaluated datasets and metrics, with statistical significance confirmed using a patient-level paired bootstrap (10,000 resamples; 95\% confidence intervals, CIs).
\begin{figure*}[!htbp]
	\centering
	\begin{subfigure}[b]{.475\columnwidth}
		\centering
		\includegraphics[width=\columnwidth]{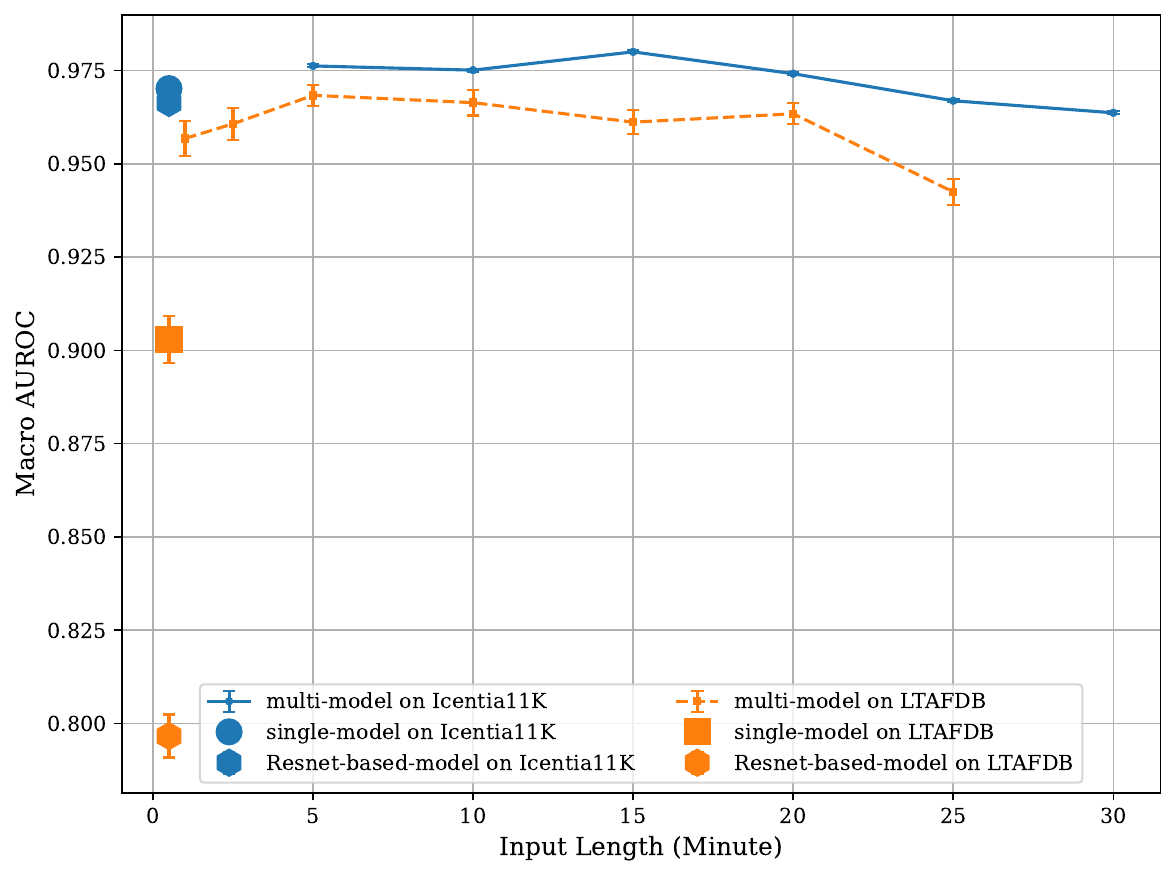}
		\caption{Performance on in-distribution datasets}
		\label{fig:id}
	\end{subfigure}
	\begin{subfigure}[b]{.475\columnwidth}
		\centering
		\includegraphics[width=\columnwidth]{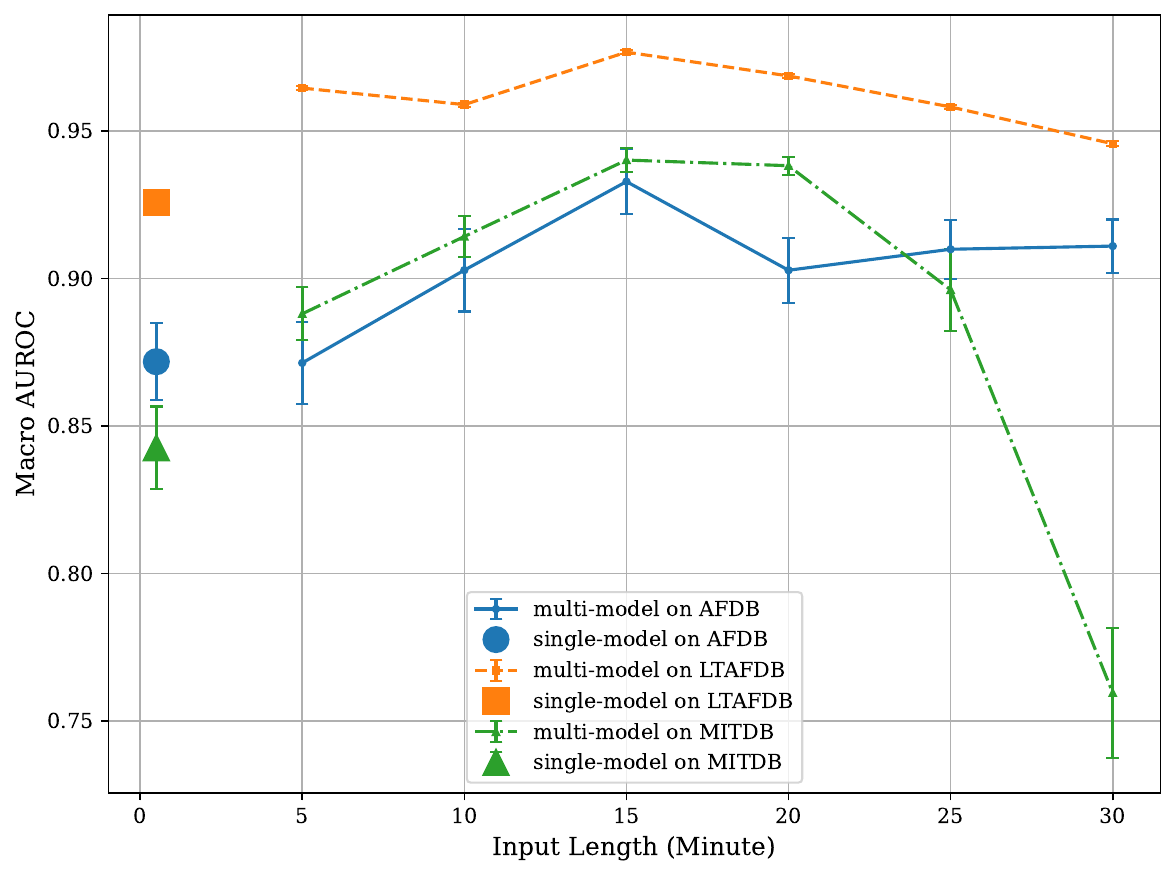}
		\caption{Performance on out-of-distribution datasets}
		\label{fig:ood}
	\end{subfigure}
	
	\caption{Performance comparison for (a) in-distribution datasets (b) out-of-distribution datasets.}
	\label{fig:results}
 \end{figure*}

\subsection{Influence of the number of input windows}

Systematic evaluation of multi-window model performance reveals clear optimal ranges for the number of input windows, with distinct patterns emerging across different training datasets and evaluation scenarios. Figure~\ref{fig:results} illustrates the performance trends across different temporal contexts. 

\heading{ID performance trends.} For models trained on Icentia11k (Table~\ref{tab:icentia_in}), the S4-based single-window baseline achieves a macro-AUROC of 0.970, establishing a high-performance baseline. The S4-based single-window model already outperforms the ResNet baseline by 0.4\% (0.9702 vs. 0.9663), confirming S4's architectural advantages before considering multi-window benefits in line with previous investigations \citep{al2025benchmarking}. Multi-window models demonstrate consistent improvements, with optimal performance achieved at 30 input windows (macro-AUROC: 0.9800, +1.0\% improvement). Uncertainty estimates are on the order of $10^{-4}$ across models. This represents improvements across all rhythm classes. At a fixed AF sensitivity of 0.9, specificity improves from 0.9033 (single-window S4) to 0.9869 (30 windows).

For LTAFDB training (Table~\ref{tab:itafdb_id}), the multi-window advantage is even more pronounced. The single-window model achieves a modest macro-AUROC of 0.9029, representing a 4.7\% improvement over the ResNet baseline (0.8621), which again confirms S4's superiority for this challenging dataset before any multi-window modeling. Multi-window models show substantial improvements from as few as two input windows. Peak performance occurs at 10 input windows (macro-AUROC: 0.9684, +7.3\% improvement), with exceptional gains in atrial fibrillation detection (AF: 0.8319 to 0.9841, +18.3\%). Notably, the performance remains consistently high across the 10-40 window range (macro-AUROC: 0.9612-0.9684), with different rhythm classes achieving their optimal performance at different points within this stable range: normal rhythm (N) at 20 windows (0.9950), and SVTA at 40 windows (0.9285). Correspondingly, AF specificity rises from 0.9794 (single-window S4) to 0.9983 (30 windows). Among these models, the 20-window configuration achieves performance that is statistically equivalent to the best model (macro-AUROC: 0.9664), demonstrating the robustness of the multi-window approach across this temporal range and providing flexibility in clinical deployment scenarios.

\begin{table}[!h]
    \centering
    \caption{Icentia11k in-distribution performance (model trained and evaluated on Icentia11k). ResNet baseline included to validate S4 architectural choice. \underline{\textbf{Underlined bold macro-AUROC}}: best performing model. \textbf{Bold}: highest values within each class. Spec.: AF specificity at sensitivity of 0.9.}
    \label{tab:icentia_in}
\resizebox{\linewidth}{!}{
\begin{tabular}{ccccccc}
\toprule
\multicolumn{1}{l}{\multirow{2}{*}{\textbf{\makecell{Model\\Type}}}} & \multicolumn{1}{l}{\multirow{2}{*}{\textbf{\makecell{Input\\Windows}}}} & \multicolumn{4}{c}{\textbf{AUROC}} & \multicolumn{1}{c}{\textbf{Spec.}}                          \\ \cline{3-6} \cline{7-7}
\multicolumn{1}{l}{}                                    & \multicolumn{1}{l}{}                  & \textbf{Macro} & \textbf{AF} & \textbf{AFLT} & \textbf{N} & \textbf{AF}       \\ \midrule
\multicolumn{1}{l}{\makecell{Single-window\\Model(ResNet)}}      & 1                                     & 0.9663$\pm$0.0001          & 0.9846         & 0.9711          & 0.9430 & 0.9008          \\
\multicolumn{1}{l}{\makecell{Single-window\\Model(S4)}}      & 1                                     & 0.9702$\pm$0.0001          & 0.9931          & 0.9665          & 0.9510 & 0.9033          \\ \midrule
\multirow{6}{*}{\makecell{Multi-window\\Model (S4ECG)}}          & 10                                    & 0.9762$\pm$0.0001          & 0.9953          & 0.9764          & 0.9570 & 0.9048          \\
                                                        & 20                                    & 0.9751$\pm$0.0002          & \textbf{0.9948} & 0.9746          & 0.9559 & 0.9645          \\
                                                        & 30                                    & \underline{\textbf{0.9800$\pm$0.0001}} & 0.9944          & \textbf{0.9811} & \textbf{0.9645} & \textbf{0.9869}          \\
                                                        & 40                                    & 0.9742$\pm$0.0001          & 0.9936          & 0.9717          & 0.9572 & 0.9607 \\
                                                        & 50                                    & 0.9669$\pm$0.0001          & 0.9947          & 0.9644          & 0.9416 & 0.9771          \\
                                                        & 60                                    & 0.9637$\pm$0.0002          & 0.9943          & 0.9630          & 0.9337 & 0.9644          \\ \bottomrule
\end{tabular}}
\end{table}

\begin{table}[!h]
	\centering
    \caption{LTAFDB in-distribution performance (model trained and evaluated on LTAFDB): Confirming S4's superiority over CNN (ResNet) established in Table~\ref{tab:icentia_in}, and combined with prior evidence of S4's advantages over LSTM and Transformers~\citep{wang2025s4sleep}, justifying our focus on S4-based architectures. \underline{\textbf{Underlined bold macro-AUROC}}: best performing model. \textbf{Bold macro-AUROC}: statistically equivalent to best model. \textbf{Bold class-AUROC}: highest values within each class. Spec.: AF specificity at sensitivity of 0.9.}
    \label{tab:itafdb_id}
\resizebox{\linewidth}{!}{
\begin{tabular}{ccccccc}
\toprule
\multicolumn{1}{l}{\multirow{2}{*}{\textbf{\makecell{Model\\Type}}}} & \multicolumn{1}{l}{\multirow{2}{*}{\textbf{\makecell{Input\\Windows}}}} & \multicolumn{4}{c}{\textbf{AUROC}} & \multicolumn{1}{c}{\textbf{Spec.}}                          \\ \cline{3-6} \cline{7-7}
\multicolumn{1}{l}{}                                     & \multicolumn{1}{l}{}                   & \textbf{Macro} & \textbf{AF} & \textbf{N}     & \textbf{SVTA} & \textbf{AF} \\ \midrule
\multicolumn{1}{l}{\makecell{Single-window\\Model(ResNet)}}       & 1                                      & 0.8621$\pm$0.0058         & 0.8574         & 0.9748          & 0.7542 & 0.9564         \\
\multicolumn{1}{l}{\makecell{Single-window\\Model(S4)}}       & 1                                      & 0.9029$\pm$0.0062         & 0.8319         & 0.9868          & 0.8899 & 0.9794         \\ \midrule
\multirow{8}{*}{\makecell{Multi-window\\Model(S4ECG)}}           & 2                                      & 0.9568$\pm$0.0047          & \textbf{0.9909}& 0.9883          & 0.8912 & 0.9897         \\
                                                         & 5                                      & 0.9607$\pm$0.0043          & 0.9872         & 0.9915          & 0.9035 & 0.9935         \\
                                                         & 10                                     & \underline{\textbf{0.9684$\pm$0.0029}} & 0.9841         & 0.9941          & 0.9269 & 0.9936         \\
                                                         & 20                                     & \textbf{0.9664$\pm$0.0034} & 0.9844         & \textbf{0.9950} & 0.9198 & 0.9960         \\
                                                         & 30                                     & 0.9612$\pm$0.0032          & 0.9711         & 0.9914          & 0.9210 & \textbf{0.9983}         \\
                                                         & 40                                     & 0.9634$\pm$0.0029          & 0.9710         & 0.9908          & \textbf{0.9285} & 0.9925 \\
                                                         & 50                                     & 0.9425$\pm$0.0036          & 0.9782         & 0.9921          & 0.8571 & 0.9969         \\
                                                         & 60                                     & 0.9470$\pm$0.0035          & 0.9751         & 0.9929          & 0.8730 & 0.9910         \\ \bottomrule
\end{tabular}}
\end{table}

\heading{Class-specific analysis} The rhythm-specific improvements reveal important insights into the clinical value of temporal context. AF detection shows the most consistent improvements, with multi-window models often reaching AUROC $\ge 0.98$ in ID settings and $\ge 0.95$ in OOD evaluations. This finding aligns with the clinical understanding that atrial fibrillation episodes often exhibit characteristic temporal patterns that extend beyond individual 30-second windows. In contrast, normal rhythm classification shows more modest but consistent improvements, suggesting that the temporal context helps distinguish true normal rhythms from transient artifacts or brief arrhythmic episodes.

\heading{Moderate sequence length advantages} As illustrated in Figure~\ref{fig:id}, the performance trajectory is non-monotonic, characterized by an initial rapid improvement followed by a plateau. A slight performance degradation is observed beyond 40 windows. \added{We hypothesize that this decline primarily reflects optimization challenges than architectural limitations: although S4 layers can theoretically capture long-range dependencies with linear complexity, longer sequences induce a more complex loss landscape with numerous local minima, making gradient-based optimization more difficult and sensitive to initialization. This interpretation is supported by prior work on sleep staging}~\citep{wang2024assessing}, \added{which demonstrated that curriculum training—progressively extending the sequence length during training—recovers the performance lost under direct long-sequence training, suggesting that the limitation lies in the optimization trajectory rather than in the model architecture or data characteristics. Secondary contributing factors may include diminishing diagnostic returns beyond the physiologically meaningful timescale of approximately 10--20 minutes for most arrhythmias. Taken together, these effects suggest that moderate input sequence lengths (20--40 windows) capture diagnostically relevant dependencies while avoiding the optimization difficulties associated with very long sequences.}  A qualitatively similar effect was observed in a recent study~\citep{Mehari2023S4} upon studying the optimal input size for interpreting 10-second 12-lead ECGs, which suggested that these signals do not carry long-range interactions beyond 2.5-3 seconds. While the latter can be assumed to be stationary across 10 seconds, the results achieved in this work suggest that diagnostically relevant long-range interactions for arrhythmia detection in non-stationary long-term ECGs remain limited to time frames around 10-15 minutes.

\subsection{OOD evaluation}

\begin{table}[!h]
    \centering
    \caption{Out-of-distribution evaluation (model trained on Icentia11k, evaluated on AFDB, MITDB, and LTAFDB). \underline{\textbf{Underlined bold macro-AUROC}}: best performing model. \textbf{Bold macro-AUROC}: statistically equivalent to best model. \textbf{Bold class-AUROC}: highest values within each class. Spec.: AF specificity at sensitivity of 0.9.}
    \label{tab:ood}
\resizebox{\columnwidth}{!}{
\begin{tabular}{cccccccc}
\toprule
\multirow{2}{*}{\textbf{Dataset}} & \multirow{2}{*}{\textbf{\makecell{Model\\Type}}} & \multirow{2}{*}{\textbf{\makecell{Input\\Windows}}} & \multicolumn{4}{c}{\textbf{AUROC}} & \multicolumn{1}{c}{\textbf{Spec.}} \\ \cline{4-7} \cline{8-8}
& & & \textbf{Macro} & \textbf{AF} & \textbf{AFLT} & \textbf{N} & \textbf{AF} \\ 
\midrule
\multirow{7}{*}{\textbf{AFDB}} 
& \makecell{Single-window\\Model} & 1  & 0.8718$\pm$0.0127 & 0.9896 & 0.6899 & 0.9357 & 0.9572 \\ \cmidrule(l){2-8}
& \multirow{6}{*}{\makecell{Multi-window\\Model(S4ECG)}} 
                                & 10 & 0.8713$\pm$0.0139 & 0.8789 & 0.7637 & 0.9713 & 0.9971 \\
                                & & 20 & 0.9028$\pm$0.0135 & 0.9826 & 0.7405 & 0.9853 & 0.9988 \\
                                & & 30 & \underline{\textbf{0.9328$\pm$0.0107}} & \textbf{0.9924} & \textbf{0.8190} & \textbf{0.9870} & \textbf{0.9998} \\
                                & & 40 & 0.9028$\pm$0.0108 & 0.9513 & 0.8000 & 0.9570 & 0.9936 \\
                                & & 50 & 0.9099$\pm$0.0102 & 0.9386 & 0.8396 & 0.9515 & 0.9967 \\
                                & & 60 & 0.9109$\pm$0.0095 & 0.9207 & 0.8765 & 0.9357 & 0.9517 \\ 
\midrule
\multirow{7}{*}{\textbf{MITDB}} 
& \makecell{Single-window\\Model} & 1  & 0.8426$\pm$0.0140 & 0.9259 & 0.9035 & 0.6983 & 0.7962 \\ \cmidrule(l){2-8}
& \multirow{6}{*}{\makecell{Multi-window\\Model(S4ECG)}} 
                                & 10 & 0.8880$\pm$0.0127 & 0.9345 & 0.9456 & 0.7838 & 0.8626 \\
                                & & 20 & 0.9142$\pm$0.0091 & 0.9500 & 0.9878 & 0.8048 & 0.8191 \\
                                & & 30 & \underline{\textbf{0.9401$\pm$0.0074}} & \textbf{0.9845} & \textbf{0.9893} & 0.8465 & 0.8226 \\
                                & & 40 & \textbf{0.9382$\pm$0.0035} & 0.9615 & 0.9774 & \textbf{0.8757} & \textbf{0.9743} \\
                                & & 50 & 0.8961$\pm$0.0141 & 0.9211 & 0.9434 & 0.8238 & 0.7931 \\
                                & & 60 & 0.7595$\pm$0.0224 & 0.8424 & 0.7025 & 0.7336 & 0.5161 \\ 
\midrule
\multirow{7}{*}{\textbf{LTAFDB}} 
& \makecell{Single-window\\Model} & 1  & 0.9256$\pm$0.0008 & 0.9427 & -- & 0.9085 & 0.718 \\ \cmidrule(l){2-8}
& \multirow{6}{*}{\makecell{Multi-window\\Model(S4ECG)}} 
                                & 10 & 0.9645$\pm$0.0006 & 0.9464 & -- & \textbf{0.9826} & 0.9792 \\
                                & & 20 & 0.9589$\pm$0.0007 & 0.9598 & -- & 0.9580 & \textbf{0.9884} \\
                                & & 30 & \underline{\textbf{0.9767$\pm$0.0004}} & \textbf{0.9748} & -- & 0.9786 & 0.9672 \\
                                & & 40 & 0.9686$\pm$0.0005 & 0.9608 & -- & 0.9765 & 0.9503 \\
                                & & 50 & 0.9581$\pm$0.0007 & 0.9484 & -- & 0.9678 & 0.9053 \\
                                & & 60 & 0.9456$\pm$0.0008 & 0.9399 & -- & 0.9514 & 0.9628 \\ 
\bottomrule
\end{tabular}}
\end{table}

The OOD evaluation on three external datasets provides crucial insights into model generalization capabilities and reveals that multi-window models demonstrate superior robustness across diverse clinical scenarios and acquisition protocols.

\heading{Cross-dataset generalization from Icentia11k} Models trained on the large-scale Icentia11k dataset exhibit remarkable robustness when evaluated on external datasets (AFDB, MITDB, and LTAFDB) as shown in Table~\ref{tab:ood}. For the AFDB evaluation, single-window models achieve a macro-AUROC of 0.8718, while multi-window models reach peak performance at 30 input windows (macro-AUROC: 0.9328, +7.0\% improvement). Correspondingly, AF specificity at sensitivity 0.9 improves strongly from 0.9572 (single-window) to 0.9998 (30 windows), demonstrating exceptional reduction in false positives. The improvement is particularly striking for atrial flutter detection (AFLT: 0.6899 to 0.8190, +18.7\%), demonstrating the value of temporal context for detecting this challenging arrhythmia type in OOD settings.

Similarly, on MITDB evaluation, multi-window models show consistent improvements, with optimal performance again at 30 input windows (macro-AUROC: 0.9401 vs. 0.8426 for single-window, +11.6\% improvement) and the 40-window configuration achieves statistically equivalent performance (macro-AUROC: 0.9382), demonstrating robustness across this temporal range. The AF detection specificity shows substantial improvement from 0.7962 (single-window) to 0.9743 (40 windows), reflecting enhanced clinical utility through reduced false alarms. The substantial improvement in normal rhythm classification (N: 0.6983 to 0.8465, +21.2\%) suggests that multi-window models are particularly effective at maintaining specificity in challenging OOD scenarios where signal characteristics may differ significantly from training data.

Cross-dataset evaluation from Icentia11k training to LTAFDB testing reveals robust generalization capabilities, with the 30-window model achieving a macro-AUROC of 0.9767 (+5.5\% over single-window). Most notably, AF specificity improves from 0.718 (single-window) to 0.9884 (20 windows), representing a dramatic 37.6\% increase that substantially reduces false positive burden. AF detection shows improvement from the single-window baseline (AF: 0.9427 to 0.9748, +3.4\%), demonstrating that the temporal patterns learned from Icentia11k's diverse population effectively generalize to detect atrial fibrillation in LTAFDB's AF-focused dataset. The consistently high performance across the multi-window range mirrors the stability patterns observed in the ID evaluation, suggesting that the optimal temporal window characteristics are robust across different dataset domains and patient populations.

\subsection{Quantitative performance analysis}

\heading{Performance stability across window counts} The multi-window advantage is consistently substantial across all evaluation scenarios. In-distribution improvements range from 1.0\% (Icentia11k) to 7.3\% (LTAFDB) in macro-AUROC, while OOD improvements are even more pronounced, ranging from 5.5\% to 11.6\%. These improvements represent clinically meaningful enhancements in diagnostic accuracy, particularly for rare but critical arrhythmias. Analysis of performance across different window counts reveals remarkable stability in the 20-40 window range, with peak performance consistently achieved at 30 windows for Icentia11k-trained model ID and OOD validation. This stability suggests robust optimal hyperparameter selection and practical deployment considerations, as moderate variations in sequence length do not dramatically impact performance. The observed performance curve, characterized by rapid improvements from single-window to moderate multi-window models followed by gradual degradation at excessive sequence lengths, corroborates the theoretical framework proposed by Wang and Strodthoff~\citep{wang2024assessing} regarding the optimal temporal window for physiological signal analysis. This pattern reflects the fundamental trade-off between capturing meaningful long-range dependencies and avoiding the curse of dimensionality in sequence modeling.

\heading{Rhythm-specific insights} Atrial fibrillation detection shows the most consistent improvements across all scenarios, with multi-window models often achieving AUROC $\ge 0.98$ in ID settings and $\ge 0.95$ in OOD evaluations. This finding supports the clinical intuition that atrial fibrillation episodes exhibit characteristic temporal signatures that extend beyond individual windows. Normal rhythm classification improvements, while more modest (typically 1-3\%), are consistently present and particularly valuable for maintaining specificity in clinical applications.

\heading{Cross-domain validation of moderate sequence lengths} Our results provide strong empirical validation of the theoretical insights from the sleep staging domain~\citep{wang2024assessing}, demonstrating that the principle of moderate sequence length optimization generalizes across different physiological monitoring applications. The consistent 20-40 window optimal range observed in our ECG arrhythmia detection task mirrors the findings in sleep stage classification, where similar moderate temporal windows proved most effective. This cross-domain consistency suggests a fundamental characteristic of physiological time series analysis, where intermediate-length sequences provide the optimal balance between temporal context richness and model generalization capability. Such findings have broader implications for the design of temporal models in biomedical signal processing, supporting the adoption of moderate sequence lengths as a general principle rather than a task-specific optimization.

\subsection{Qualitative insights and clinical impact}

\begin{figure*}[!htbp]
    \centering
    \includegraphics[width=\linewidth]{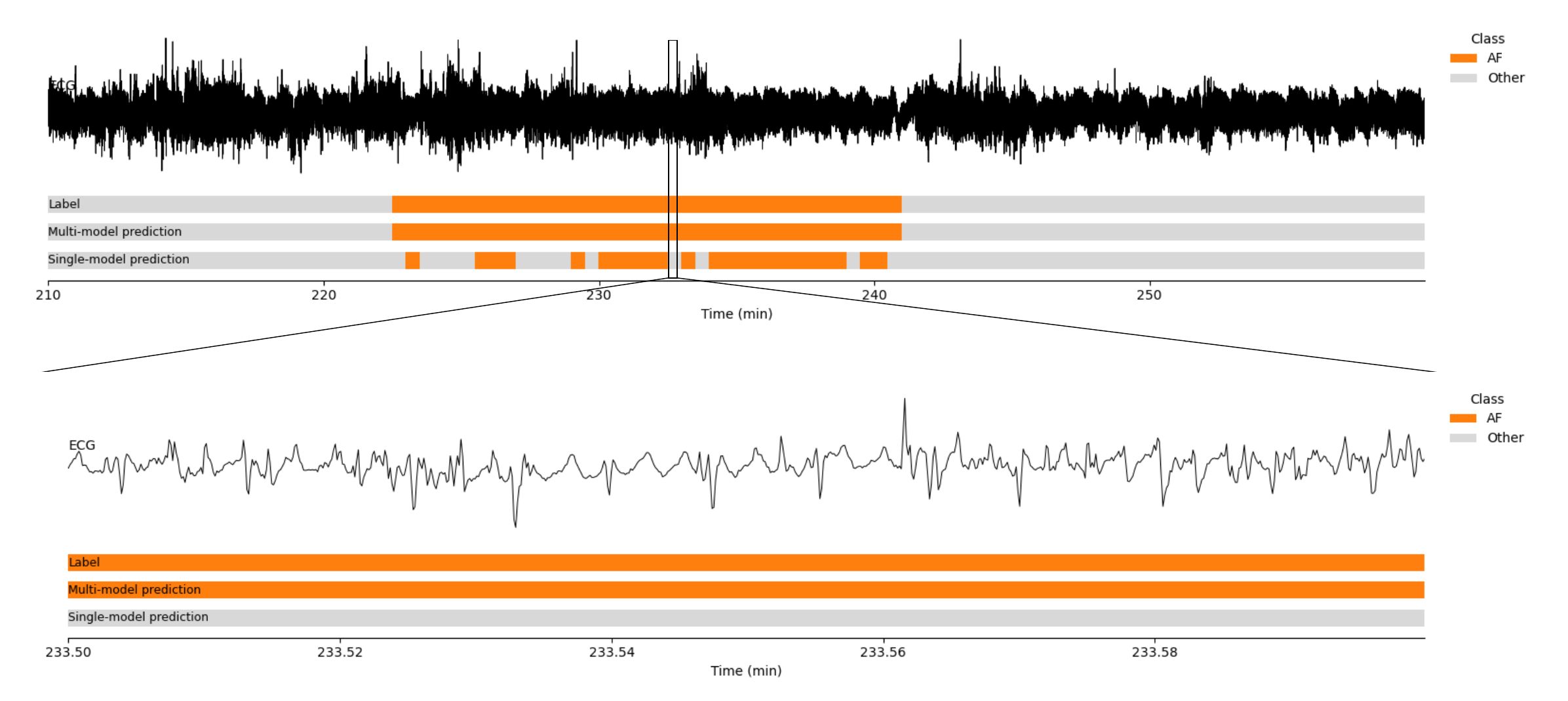}
    \caption{Qualitative comparison of model predictions on a continuous atrial fibrillation episode from LTAFDB. Top to bottom: ground truth annotation, multi-window model (30-window input), single-window model, and enlarged detail. The multi-window model maintains temporal coherence matching the ground truth, while the single-window model produces fragmented predictions with spurious interruptions, demonstrating the superior temporal consistency of the multi-window approach for arrhythmia burden estimation.}
    \label{fig:band}
\end{figure*}

To provide deeper insights into multi-window model behavior and temporal pattern recognition capabilities, we present qualitative analysis of individual ECG recordings showing how the S4ECG model, which is trained on Icentia11k at 30 windows input length, processes extended temporal sequences. Figure~\ref{fig:band} presents a representative example from a test recording in the LTAFDB dataset, illustrating the multi-window model's superior temporal coherence and arrhythmia burden estimation capabilities.

The visualization includes analysis of AF burden, comparing actual and predicted AF loads across the extended monitoring period from this LTAFDB recording. The multi-window model demonstrates superior accuracy in estimating the overall proportion of time spent in AF, which is a clinically important metric for patient risk stratification and treatment decisions.

Unlike single-window models that may produce inconsistent predictions across adjacent time segments, the multi-window S4ECG model generates temporally coherent predictions that better align with the underlying physiological patterns. The extended temporal context enables the model to maintain consistency across rhythm transitions and reduces fragmented episode detection, as demonstrated in this LTAFDB test recording.

\section{Discussion}
\label{sec:discussion}
\subsection{Technical implications}
\heading{Summary} This work presents, to our knowledge, the first systematic, cross-dataset investigation of multi-window deep learning approaches for ECG arrhythmia detection, introducing S4ECG, a novel architecture that leverages structured state space models to capture long-range temporal dependencies in cardiac rhythm analysis. Our systematic evaluation across four major ECG databases demonstrates consistent and substantial improvements of multi-window models over conventional single-window approaches, with particularly notable gains in out-of-distribution scenarios.

\heading{Multi-window paradigm} The most significant finding is the consistent optimal performance achieved in the 20-40 window range (10-20 minutes of ECG data), suggesting fundamental characteristics of cardiac rhythm analysis that extend beyond dataset-specific artifacts. Multi-window models achieve statistically significant improvements ranging from 1.0\% to 11.6\% in macro-AUROC across different evaluation scenarios, with particularly striking gains for challenging arrhythmia types such as atrial flutter (+18.7\% improvement in OOD settings). Similarly, multi-window models, in comparison to single-window models, show a substantially increased specificity at a fixed sensitivity level of 0.9, indicating a substantial reduction in false positive predictions. The stability of these improvements across diverse datasets, acquisition protocols, and patient populations provides strong evidence for the generalizability of multi-window approaches. 

We envision that this work will contribute to a methodological transition in arrhythmia detection algorithm design, encompassing both temporal modeling evolution—from single-window analysis toward temporally-aware multi-window approaches—and architectural advancement from traditional CNN/LSTM frameworks toward efficient structured state space models that better align with clinical practice and the inherent temporal nature of cardiac arrhythmias.

\heading{Broader impact} Beyond the empirical findings, this work establishes a methodological framework for investigating temporal dependencies in physiological time series analysis. The consistent validation of moderate sequence length principles across ECG arrhythmia detection and sleep staging domains suggests broader applicability of these design principles in biomedical signal processing. The S4ECG architecture provides a computationally efficient solution for long-range dependency modeling that scales linearly with sequence length.

\subsection{Clinical implications}
\added{Our findings have several clinical implications.} \deleted{Our findings demonstrate substantial clinical implications that address critical gaps in contemporary arrhythmia monitoring and management.} The demonstration that moderate temporal windows (10-20 minutes) yield optimal performance fundamentally challenges conventional arrhythmia detection paradigms. 

Furthermore, the superior out-of-distribution performance of multi-window models indicates a high degree of robustness—a critical requirement for real-world deployment, which was explicitly acknowledged for example in the FDA's Software as a Medical Device (SaMD) guideline~\citep{FDA2017SaMD} and the action plan on AI/ML-based SaMD~\citep{FDA2021AIActionPlan}. This robustness addresses the well-documented challenge of domain shift in medical AI applications, where models frequently underperform when applied to patient populations, recording devices, or clinical environments that differ from training conditions. \deleted{The enhanced generalizability of our approach suggests improved performance across diverse healthcare settings, patient demographics, and ECG acquisition systems without requiring extensive retraining or calibration. }More specifically, we anticipate impact on the following use cases:

\heading{Enhanced diagnostic precision and clinical burden reduction}
Improved overall predictive performance at fixed sensitivity can translate into higher specificity, reducing false positives, unnecessary clinical interventions, costs, and patient anxiety. False positive rates have been reported to be high in certain conventional automated systems~\citep{bumgarner2018smartwatch,tison2018passive}, contributing to increased emergency department utilization from consumer-grade devices~\citep{seshadri2020accuracy,rajakariar2020accuracy}. As a result, the proposed approach can improve patient experience and resource utilization without compromising diagnostic safety.

\heading{Advanced arrhythmia characterization and temporal dynamics} 
Understanding the temporal dynamics characteristics of arrhythmia is key for more fine-grained understanding of arrhythmia subclasses, such as paroxysmal and persistent in the case of atrial fibrillation, and is believed to lead to clinically actionable insights into disease progression and treatment response~\citep{charitos2014clinical, de2010progression}. The improved accuracy but also temporal consistency, see Figure~\ref{fig:band}, of the proposed approach aligns with this goal. 

\heading{Atrial fibrillation burden assessment} Precise AF burden quantification is increasingly recognized as a critical determinant of stroke risk and therapeutic decision-making~\citep{chen2018atrial}. Studies demonstrate that even modest AF burden ($>$0.5\%) correlates with increased thromboembolic risk, emphasizing the clinical importance of accurate measurement. Our approach enables continuous, high-resolution burden assessment that can inform both acute and chronic management strategies.

\heading{Temporal pattern recognition} The detection of changes in arrhythmia patterns over time provides insights into disease progression and treatment efficacy. Circadian variations in arrhythmia occurrence can reveal underlying triggers, with nocturnal episodes often associated with sleep-disordered breathing and diurnal episodes linked to sympathetic activation~\citep{charitos2014clinical,de2010progression}. Such pattern recognition facilitates targeted therapeutic interventions and lifestyle modifications. 

\heading{Paroxysmal episode detection} The identification of short paroxysms, particularly those lasting seconds to minutes, addresses a significant limitation of conventional monitoring systems. These brief episodes, often asymptomatic, may nonetheless contribute to stroke risk in patients with cryptogenic cerebrovascular events~\citep{sanna2014cryptogenic,gladstone2014atrial}. Enhanced sensitivity for paroxysmal detection is particularly relevant for post-ablation monitoring, where early recurrence detection during the blanking period can inform subsequent management strategies.

\heading{Transition state analysis} The recognition of brief interruptions or transitions within arrhythmic episodes provides mechanistic insights into arrhythmia maintenance and termination~\citep{nattel2017controversies}. Analysis of onset and termination patterns can inform catheter ablation strategies by identifying critical regions for intervention. Additionally, the detection of mode switching between different arrhythmic patterns (e.g., atrial fibrillation to atrial flutter) can reveal information about the underlying electrophysiological substrate.

\deleted{\heading{Precision medicine and individualized treatment strategies} These enhanced diagnostic capabilities represent significant advancement toward precision electrophysiology, where treatment strategies are tailored to individual arrhythmia characteristics, patient physiology, and response patterns}
\deleted{The integration of high-resolution temporal analysis with clinical risk factors enables more sophisticated risk stratification algorithms that account for both arrhythmia burden and pattern variability.}

\subsection{Limitations of the study}

While our evaluation encompasses multiple databases and scenarios, several limitations should be acknowledged. First, our study focuses exclusively on supervised learning paradigms, which require extensive labeled data that may not always be available in clinical settings. The reliance on expert-annotated rhythm labels limits scalability to larger, unlabeled ECG datasets that are increasingly common in clinical practice. Nevertheless, the multi-window S4ECG design is naturally compatible with self-supervised objectives that exploit inter-window temporal relations for representation learning without explicit labels; future work should explore such adaptations to harness large unlabeled datasets and further improve generalization and robustness.

Second, our evaluation uses retrospective datasets; validation in real-time clinical monitoring systems remains to be established. Although S4-based models are computationally efficient, thorough assessment in edge-computing environments typical of wearable and mobile health devices is needed.

Third, the current approach processes fixed-length sequences, which may not optimally capture variability in arrhythmic episode duration. Adaptive sequence-length mechanisms that adjust context based on rhythm stability could further enhance performance.

\section{Summary and conclusion}
\label{sec:summary}
In this work, we present S4ECG, a deep learning architecture that exploits structured state space models to capture diagnostically relevant long-range dependencies in electrocardiogram signals. Through systematic evaluation across four major ECG databases, we demonstrate that the joint analysis of multiple consecutive windows yields substantial improvements over conventional single-window approaches, particularly in challenging out-of-distribution scenarios. Crucially, our findings identify an optimal temporal context of 10--20 minutes, reflecting fundamental physiological timescales of cardiac rhythm dynamics that transcend dataset-specific characteristics. By significantly enhancing robustness and reducing false positive rates, this approach addresses key barriers to the reliability of automated arrhythmia monitoring. These results advocate for a methodological transition in algorithm design—moving beyond isolated morphological analysis toward efficient, temporally-aware sequence modeling that aligns with the continuous nature of cardiac pathophysiology.

\section*{Data availability statement}
All datasets used in this study are publicly available through PhysioNet (\url{https://physionet.org}): Icentia11k, LTAFDB, AFDB, and MITDB. The source code implementation as well as the training-validation-test data splits are available at the accompanying repository~\citep{repo}.



\bibliographystyle{plainnat} 
\bibliography{bibfile}



\newif\ifdraft
\drafttrue
\newif\ifstandalone
\standalonefalse

\ifstandalone
\PassOptionsToPackage{table}{xcolor}

\documentclass[lettersize,journal]{IEEEtran}

\usepackage{xcolor}  
\usepackage[superscript]{citep}
\usepackage{amsmath,amssymb,amsfonts}
\usepackage{bbm}
\usepackage{graphicx}

\usepackage{hyperref}
\usepackage{multirow}
\usepackage{cleveref}
\usepackage{comment}
\usepackage{hhline}
\usepackage{todonotes}
\usepackage{caption,subcaption}

\usepackage{siunitx}
\sisetup{round-mode=places,round-precision=3}
\usepackage{booktabs}
\usepackage{csvsimple}

\usepackage{datatool}
\usepackage{array}
\usepackage{placeins}
\usepackage{float}
\usepackage{pifont} 
\usepackage{makecell}

\definecolor{improvement}{RGB}{144,238,144} 
\definecolor{degradation}{RGB}{255,182,193} 
\definecolor{nochange}{RGB}{255,255,255}

\newcommand\colnst[1]{{\color{red}#1}}
\newcommand\coltw[1]{{\color{blue}#1}}

\newcommand{\heading}[1]{\noindent\textbf{#1}}

\newcommand{\diffcellsplitside}[4]{%
    \makecell[l]{%
        \colorbox{#1}{\parbox{0.7cm}{\centering \scriptsize #2}} \hspace{0.2cm}%
        \colorbox{#3}{\parbox{0.7cm}{\centering \scriptsize #4}}%
    }
}
\usepackage[normalem]{ulem} 
\newcommand{\stkout}[1]{\ifmmode\text{\sout{\ensuremath{#1}}}\else\sout{#1}\fi}

\newcommand{\citepx}[1]{\mbox{\citep{#1}}}
\newcommand{\citeppx}[1]{\mbox{\citepp{#1}}}
\newcommand{\citeptx}[1]{\mbox{\citept{#1}}}
\ifdraft

\newcommand{\added}[1]{\textcolor{blue}{#1}}
\newcommand{\deleted}[1]{\textcolor{red}{\stkout{#1}}}
\newcommand{\replaced}[2]{\textcolor{blue}{#1} \textcolor{red}{\stkout{#2}}}
\newcommand{\deletedfloat}[1]{}
\newcommand{\commented}[1]{\textcolor{blue}{#1}}
\else

\newcommand{\added}[1]{#1}
\newcommand{\deleted}[1]{}
\newcommand{\replaced}[2]{#1}
\newcommand{\deletedfloat}[1]{}
\newcommand{\commented}[1]{}
\fi

\newcommand\citepp[1]{\citep{#1}}
\newcommand\citept[1]{\citep{#1}}
\newcommand{\pcn}[1]{\textcolor{magenta}{#1}} 

\def\BibTeX{{\rm B\kern-.05em{\sc i\kern-.025em b}\kern-.08em
    T\kern-.1667em\lower.7ex\hbox{E}\kern-.125emX}}
\markboth{Preprint}{Wang \MakeLowercase{\textit{et al.}}: S4ECG: Exploring the impact of long-range interactions for arrhythmia prediction (Jan 2026)}

\renewcommand{\appendix}{Online Supplementary Material}

\begin{document}
\appendices
\else
\renewcommand{\thesection}{S}
\setcounter{page}{1} 
\section{Online supplementary material}

\fi

\subsection{Dataset details and preprocessing}
\label{app:datasets}
\subsubsection{Dataset details}
\heading{Icentia11k} The Icentia11k dataset~\citep{tanicentia11k,icentia11kphysionet} serves as our primary large-scale training resource, comprising continuous single-lead ECG recordings from 11,000 patients monitored over extended periods. This dataset represents the largest publicly available collection of annotated ECG data for arrhythmia research, with recordings spanning durations from several hours to multiple weeks. Each recording is sampled at 250~Hz and includes expert annotations for various cardiac rhythm types, including atrial fibrillation (AF), atrial flutter (AFL), and normal sinus rhythm (N). The dataset's substantial size and temporal extent make it particularly well-suited for investigating long-range temporal dependencies in cardiac rhythm analysis. Patient demographics span a diverse age range with balanced representation across gender groups, providing a robust foundation for model development.

\heading{Long-Term AF Database (LTAFDB)} The LTAFDB~\citep{ltafdb} serves both as a secondary training dataset and for cross-dataset OOD evaluation, featuring 84 long-term ECG recordings from patients with documented atrial fibrillation episodes. These recordings, sampled at 128~Hz, capture the natural progression and variability of atrial fibrillation patterns over extended monitoring periods ranging from 24 to 25 hours per patient. The database includes comprehensive rhythm annotations encompassing atrial fibrillation (AF), normal sinus rhythm (N), and supraventricular tachyarrhythmia (SVTA). This dataset's focus on atrial fibrillation provides complementary training data with different temporal characteristics compared to Icentia11k, while also enabling assessment of model performance across varied dataset scales and sampling frequencies in both ID and OOD scenarios.

\heading{MIT-BIH Atrial Fibrillation Database (AFDB)} For OOD evaluation, we utilize the MIT-BIH AFDB~\citep{afdb}, which contains 25 long-term ECG recordings specifically selected to include significant episodes of atrial fibrillation. These recordings, sampled at 250~Hz with durations of approximately 10 hours each, provide ground truth annotations for atrial fibrillation, atrial flutter, and normal rhythm segments. The database's careful curation and well-characterized arrhythmia episodes make it an ideal benchmark for evaluating model generalization capabilities beyond the training distribution.

\heading{MIT-BIH Arrhythmia Database (MITDB)} The MITDB~\citep{mitdb} also serves as an OOD evaluation dataset, comprising 48 half-hour excerpts of two-channel ambulatory ECG recordings from 47 subjects. These recordings, sampled at 360~Hz, include comprehensive beat-by-beat annotations for various arrhythmia types. While primarily designed for beat-level classification tasks, we adapt the annotations to window-level rhythm classification to maintain consistency with our experimental framework. The database's different sampling rate, recording duration, and patient population characteristics provide a stringent test of model robustness across diverse acquisition protocols and demographic variations.

\subsubsection{Data preprocessing}
We implement a standardized preprocessing pipeline that addresses the heterogeneity in sampling rates, signal durations, and annotation formats present in the original datasets. All ECG signals are standardized to a uniform sampling rate of 128~Hz using the first available channel, and rhythm annotations are processed to generate window-level labels through a label aggregation process that preserves the temporal distribution of rhythm types within each 30-second window. \added{ We do not perform explicit ECG denoising or baseline correction; no frequency-selective filtering is applied.}

\heading{Signal preprocessing} The ECG signals are processed directly from the original PhysioNet format without additional normalization, as the datasets already provide calibrated recordings in millivolts (mV). This preserves the original signal characteristics and amplitude relationships as intended by the dataset creators. Given the varying native sampling rates across datasets, we standardize all ECG signals to a uniform sampling rate of 128~Hz using the resampy library, which employs high-quality resampling with automatic anti-aliasing. For LTAFDB, which already has a native sampling rate of 128~Hz, no resampling is performed. This standardization ensures computational efficiency while preserving the clinically relevant frequency components of ECG signals, which typically contain most diagnostic information below 50~Hz.

\heading{Rhythm annotation processing} We extract rhythm annotations from the PhysioNet annotation files (.atr) by identifying rhythm change markers (annotated with "+" symbols in the original datasets). These rhythm annotations define temporal segments with consistent rhythm types including normal sinus rhythm (N), atrial fibrillation (AF), atrial flutter (AFL), and supraventricular tachyarrhythmia (SVTA). Additional rhythm types such as "Unknown" or unclassified segments are preserved in the dataset but excluded from loss calculation during training to ensure model optimization focuses on well-defined rhythm categories. We generate sample-level rhythm labels that assign each time point to its corresponding rhythm class. The rhythm label segmentation mask is then resampled to match the target sampling rate of 128~Hz, ensuring temporal alignment between signal data and rhythm annotations.

\begin{figure*}[!htbp]
    \centering
    \includegraphics[width=0.95\textwidth]{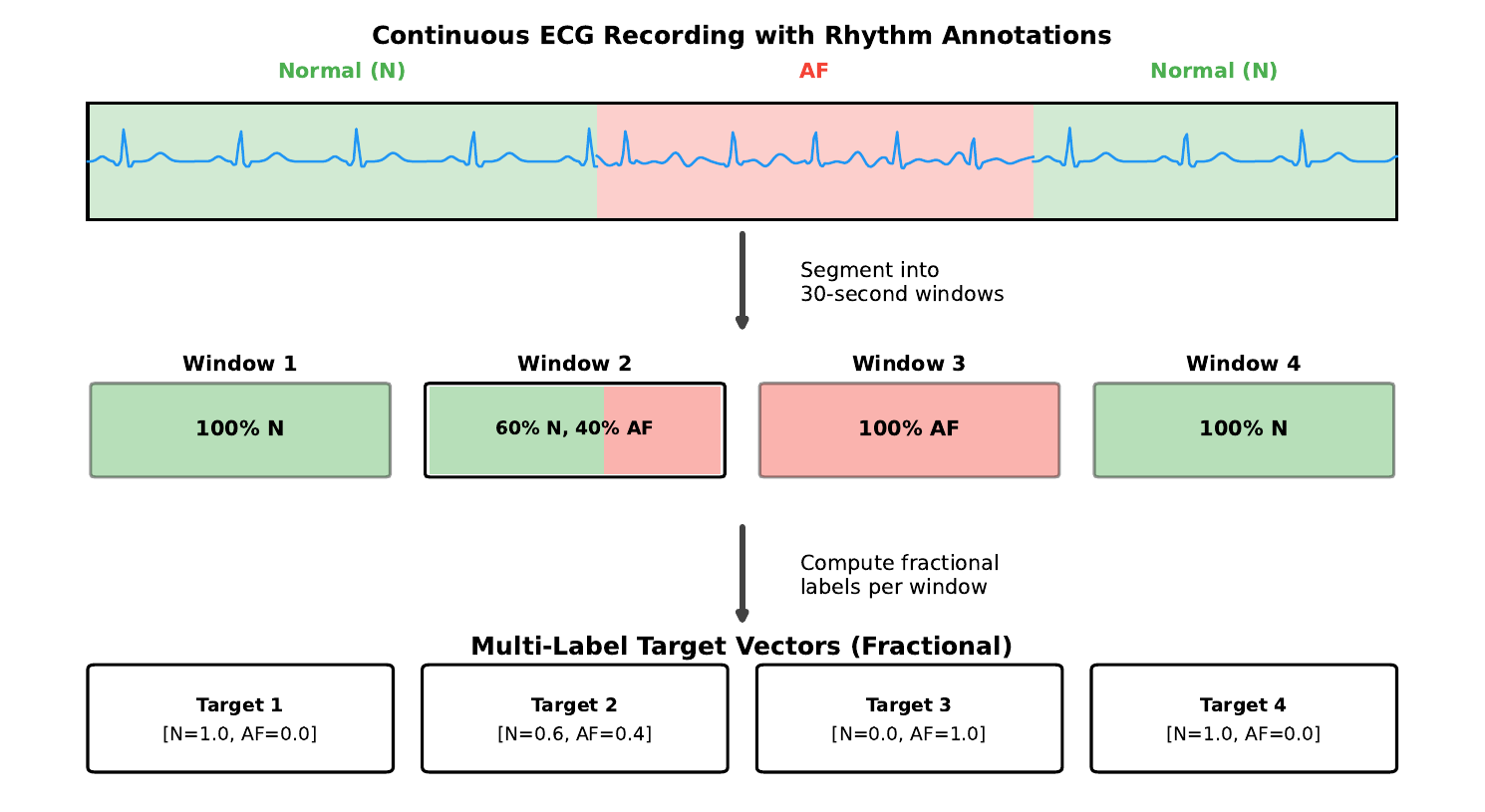}
    \caption{\added{Schematic of label generation. Continuous rhythm annotations are segmented into 30-second windows and converted to fractional multi-label targets.}}
    \label{fig:schematic}
\end{figure*}

\heading{Window-level target generation} Our implementation transforms the continuous rhythm segmentation masks into window-level labels through a label aggregation process\added{, as illustrated in Figure~\ref{fig:schematic}}. For each 30-second window (corresponding to 3,840 samples at 128~Hz), we count the number of samples belonging to each rhythm class within that temporal window. The window-level label is then computed as the fraction of time each rhythm class is present within the window, creating soft labels that preserve the temporal distribution of rhythm types. For example, if a window contains 60\% normal rhythm and 40\% atrial fibrillation samples, the resulting window label reflects these proportions rather than selecting a single predominant class. This approach maintains the rich temporal characteristics of the original PhysioNet annotations while providing window-level supervision suitable for multi-window sequence modeling. For recordings that do not divide evenly into 30-second segments, we discard the remaining partial window to maintain consistent input dimensions across all samples.

\subsubsection{Architectural details}
\label{app:repro}
\added{Dropout is applied inside S4 residual blocks. Unless stated otherwise, we follow the S4Sleep setting for S4-based models (dropout 0.2).
We use the normalization strategy of the reference S4 implementation (pre-normalization in residual blocks). Convolutional and linear layers use standard framework defaults; S4 parameters follow the reference S4 initialization. All model components (encoder, predictor, head) are optimized jointly using AdamW with a constant learning rate. No stage-wise changes of optimizer settings are used. We train separate models for each configured number of input windows $K$. During training and validation, long recordings are partitioned into consecutive, non-overlapping segments whose length matches the model input size; any trailing remainder shorter than the input size is discarded. We report  hyperparameters in Table~S\ref{tab:arch_spec}.}

\begin{table}[!t]
\centering
\caption{Model architecture specification and regularization.}
\label{tab:arch_spec}
\resizebox{\columnwidth}{!}{
\begin{tabular}{ll}
\toprule
\textbf{Item} & \textbf{Setting} \\
\midrule
Single window duration / sampling & 30\,s windows, model input at 128\,Hz (3,840 samples/window) \\
Conv front-end & 2$\times$Conv1D (128 ch), kernel=3, stride=2, activation=ReLU \\
Encoder & S4 blocks: 4 layers, bidirectional, $d{=}512$, $d_{\mathrm{state}}{=}64$ \\
Predictor & S4 blocks: 4 layers, bidirectional, $d{=}512$, $d_{\mathrm{state}}{=}64$ \\
Head & Linear \\
Dropout & S4-block dropout=0.2 \\
Normalization & Pre-norm LayerNorm in residual blocks; BatchNorm in conv front-end  \\
Initialization & Conv/linear: framework defaults; S4: reference implementation defaults \\
Loss & BCE with fractional window targets \\
Optimizer & AdamW \\
Learning rate & $1\times10^{-3}$ (constant; no scheduler) \\
Weight decay & $1\times10^{-3}$ \\
Effective batch size & 64 \\
\bottomrule
\end{tabular}}
\end{table}

\subsection{Training and evaluation details}
\label{app:setup}
\subsubsection{Training methodology}
We split datasets at the patient level for Icentia11k and LTAFDB, and at the recording level for AFDB and MITDB. This approach ensures that no patient's data appears in both training and evaluation sets, providing a more rigorous assessment of model robustness. For Icentia11k, we employ an 8:1:1 patient-level split, allocating 80\% of patients for training, 10\% for validation and model selection, and 10\% for in-distribution testing. Similarly, LTAFDB follows a 3:1:1 patient-level split (60\% training, 20\% validation, 20\% testing) to accommodate the smaller dataset size while maintaining sufficient data for each partition. \added{The dataset splits are available in the accompanying repository~\citep{repo}. Each patient is randomly assigned a fold number, and the train, validation, and test sets are partitioned according to fold numbers in ascending order, which prevents any form of information leakage.}

For the large-scale Icentia11k dataset, we use a streamlined training approach with 5 epochs, leveraging the substantial amount of training data (approximately 110,000 hours) to achieve convergence efficiently. For the smaller LTAFDB dataset with 84 patients and approximately 2,000 hours of recordings, we extend training to 150 epochs to ensure adequate learning from the limited data. 

The optimization strategy uses the AdamW optimizer with a fixed learning rate of $1 \times 10^{-3}$ and a weight decay of $1 \times 10^{-3}$ optimizing binary crossentropy as loss function with fractional targets as described above. Due to the computational demands of processing long multi-window sequences, we employ a memory-efficient training strategy with small batch sizes combined with gradient accumulation to achieve an effective batch size of 64. This approach enables training on sequences up to 60 windows (30 minutes of ECG data) while maintaining computational feasibility. Models with the best macro-AUROC performance on validation are selected for final evaluation.

\heading{Multi-window Training Configuration.} For multi-window models, the input size is adjusted based on the desired temporal context:
\begin{itemize}
\item 2 windows (1 min): input\_size = 7,680
\item 5 windows (2.5 min): input\_size = 19,200
\item 10 windows (5 min): input\_size = 38,400
\item 20 windows (10 min): input\_size = 76,800  
\item 30 windows (15 min): input\_size = 115,200
\item 40 windows (20 min): input\_size = 153,600
\item 50 windows (25 min): input\_size = 192,000
\item 60 windows (30 min): input\_size = 230,400
\end{itemize}
Each configuration maintains the fixed window length of 3,840 samples (30 seconds at 128~Hz), ensuring consistent window-level processing while varying the inter-window temporal context.

\subsection{Performance evaluation}
We evaluate model performance using a comprehensive set of metrics tailored to the multi-label nature of our rhythm classification task. \added{Since our model outputs independent probability scores $p_c \in [0,1]$ for each rhythm class $c$ via sigmoid activation, converting these to discrete class predictions depends on the evaluation context:
\begin{itemize}
\item \textbf{AUROC computation}: Continuous probability outputs are used directly without thresholding.
\item \textbf{Sensitivity/Specificity at a fixed operating point}: Per-class thresholds $\theta_c$ are tuned to achieve a specified target sensitivity (see Section~S7 for details).
\item \textbf{Confusion matrix analysis}: We take $\arg\max_c p_c$ to assign the dominant predicted class, enabling standard single-label visualization while acknowledging the underlying multi-label formulation (see Section~S8 for details).
\end{itemize}}

The primary evaluation metric is the macro-averaged area under the receiver operating characteristic curve (macro-AUROC), which provides equal weighting to all rhythm classes regardless of their prevalence in the dataset. This choice ensures that model performance on rare but clinically important arrhythmias (such as atrial flutter) receives appropriate consideration alongside more common rhythm types.

For each rhythm class $c$, we compute the AUROC by treating the prediction as a binary classification problem between class $c$ and all other classes. The macro-AUROC is then calculated as:
\begin{equation}
\text{macro-AUROC} = \frac{1}{C} \sum_{c=1}^{C} \text{AUROC}_c
\end{equation}
where $C$ is the total number of rhythm classes in the dataset.

In addition to macro-AUROC, we report class-specific AUROC values to provide detailed insights into model performance for individual rhythm types. This granular analysis is particularly important for understanding model behavior across different arrhythmia types and identifying potential areas for improvement.

For OOD evaluation, models trained and selected based on Icentia11k performance are evaluated on the complete LTAFDB, AFDB, and MITDB datasets, providing comprehensive assessment of cross-dataset generalization capabilities. This evaluation strategy ensures that model selection is performed independently of the test data, preventing any form of information leakage and providing unbiased estimates of model performance across diverse clinical scenarios.

\begin{figure}[!htbp]
    \centering
    \includegraphics[width=0.85\columnwidth]{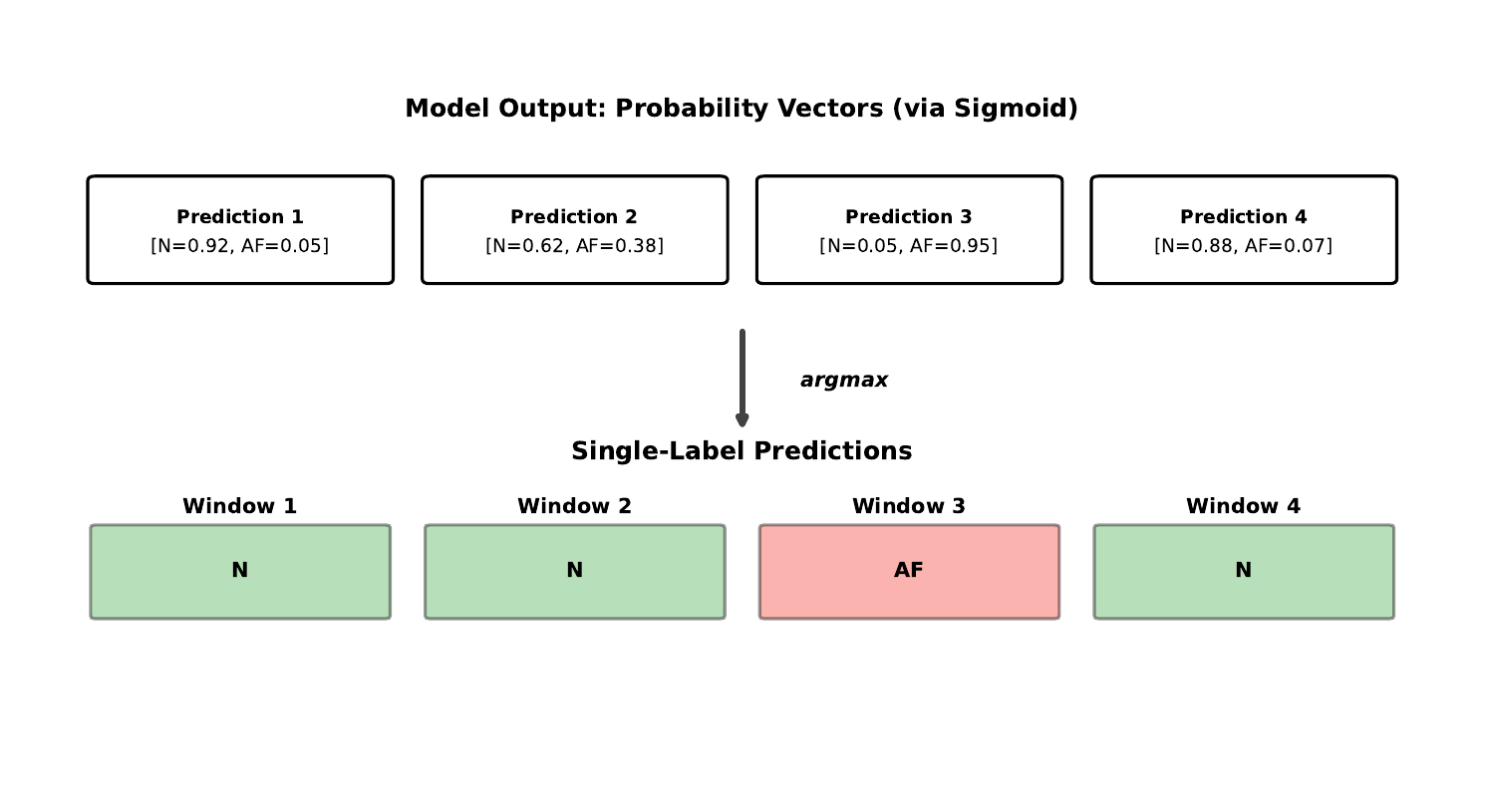}
    \caption{\added{Conversion from multi-label probability output to single-label prediction. The model outputs independent probabilities for each class via sigmoid activation. The argmax operation selects the class with highest probability as the final prediction.}}
    \label{fig:argmax}
\end{figure}

\added{Figure~\ref{fig:argmax} illustrates how the model's multi-label probability outputs are converted to single-label predictions for visualization purposes (e.g., confusion matrix analysis). The model outputs independent probability scores $p_c \in [0,1]$ for each rhythm class via sigmoid activation. To obtain a single predicted class, we apply argmax: $\hat{y} = \arg\max_c p_c$. This approach identifies the dominant predicted rhythm while acknowledging that the underlying formulation is multi-label.}

\subsection{Statistical analysis}

To assess the uncertainty of model performance metrics, we provide 95\% confidence intervals via empirical bootstrapping on the test set with 10,000 iterations. We report point estimates from evaluation on the complete test set and estimate confidence intervals using bootstrap resampling. Statistical significance between models is determined using bootstrap estimates of performance differences. If confidence intervals for the difference between the best-performing and other models do not include zero, the models are considered statistically significantly different at $\alpha = 0.05$. During bootstrapping, samples lacking positive examples for all classes are discarded and redrawn to ensure reliable macro-AUROC computation. In result tables, we report point estimates with maximal absolute deviations between point estimates and confidence interval bounds ($\pm$values).

\subsection{Computational Details}
\label{sec:appendix_computational}
The S4ECG model contains approximately 4.9 million trainable parameters, distributed across the hierarchical encoder-predictor architecture. All models were trained on NVIDIA A100 GPUs with 80~GB memory, leveraging the high-bandwidth memory and tensor core capabilities for efficient processing of long temporal sequences. The substantial memory requirements of multi-window training (particularly for 60-window sequences spanning 30 minutes of ECG data) necessitated the use of gradient accumulation strategies and memory-efficient implementations of the S4 architecture.

During training, we employ non-overlapping crops of the specified input size to ensure independent training samples and prevent data leakage between adjacent sequences. However, at inference time, we implement a sliding window approach to maximize the utilization of available temporal context and improve prediction robustness.

\subsection{Related work}

\heading{Structured state space models}
Structured state space sequence (S4) models represent a recent breakthrough in sequence modeling, offering efficient alternatives to traditional recurrent and transformer architectures for long-range dependency modeling \citep{Gu2021EfficientlyML}. S4 models leverage the mathematical framework of state space representations to capture temporal dependencies while maintaining linear computational complexity with respect to sequence length. This efficiency makes them particularly attractive for biomedical applications involving long time series.

The theoretical foundations of S4 models enable them to capture dependencies across arbitrarily long sequences without the vanishing gradient problems that plague traditional RNNs or the quadratic complexity limitations of transformer architectures. Recent work has demonstrated the effectiveness of S4 models across diverse domains, from natural language processing to time series forecasting, establishing them as a powerful tool for sequence modeling tasks.

In the context of physiological signal analysis, S4 models offer particular advantages due to their ability to capture both short-term patterns (within individual windows) and long-term dependencies (across multiple windows) within a unified framework. The hierarchical application of S4 models—at both window-level and sequence-level processing—provides a natural fit for multi-window ECG analysis, enabling efficient capture of the complex temporal dependencies inherent in cardiac rhythm patterns.

\heading{Multi-window temporal modeling}
The concept of incorporating temporal context across multiple windows has gained significant attention in biomedical signal analysis, particularly in the domain of sleep staging. The pioneering work of SeqSleepNet \citep{phan2019seqsleepnet} demonstrated that processing sequences of windows jointly rather than independently leads to substantial improvements in classification accuracy and temporal consistency. The SeqSleepNet architecture employs a hierarchical approach where individual windows are first encoded into compact representations, which are then processed by a recurrent neural network to capture inter-window dependencies.

This encoder-predictor paradigm has been further refined in subsequent work, with S4Sleep \citep{wang2025s4sleep} providing a comprehensive evaluation of different architectural components for sleep stage classification. The S4Sleep study systematically investigated the impact of various deep learning architectures, including S4 models , and established design principles for multi-window analysis in physiological time series. 

Furthermore, \citep{wang2024assessing} demonstrated that moderate sequence lengths provide optimal performance, avoiding both the limited context of single-window models and the optimization difficulties associated with excessively long sequences.

The success of multi-window approaches in sleep staging provides compelling motivation for their application to ECG analysis, given the similar temporal characteristics and physiological dependencies present in both domains. However, the direct translation of these approaches to arrhythmia detection requires careful consideration of the specific temporal patterns and clinical requirements inherent in cardiac rhythm analysis.

\subsection{Qualitative analysis thresholds}
\label{sec:appendix_qualitative}
For the qualitative analysis presented in Figure~3 of the main text, optimal classification thresholds are determined using a false negative rate-based approach that prioritizes clinical sensitivity requirements. Specifically, for each rhythm class $i$, we compute the ROC curve and select the threshold that minimizes the absolute difference between the achieved false negative rate and a clinically acceptable target rate:

\begin{equation}
\theta_i^* = \arg\min_{\theta} |(\text{FNR}(\theta) - \text{FNR}_{\text{target}})|
\end{equation}

where $\text{FNR}(\theta) = 1 - \text{TPR}(\theta)$ represents the false negative rate at threshold $\theta$. This approach is clinically motivated, as missing arrhythmic episodes (false negatives) are associated with higher clinical risk than false alarms in continuous monitoring scenarios~\citep{zweig1993receiver,fuster2006acc}. The method ensures that the model achieves the desired sensitivity level for detecting critical arrhythmic events, which is particularly important for life-threatening arrhythmias such as atrial fibrillation where early detection is crucial for preventing stroke and other complications.

For this study, we set $\text{FNR}_{\text{target}} = 0.1$ (corresponding to 90\% sensitivity) representing a high-sensitivity operating point suitable for arrhythmia screening scenarios where minimizing false negatives is prioritized.

\added{\subsection{Confusion Matrix Analysis}}
\label{sec:appendix_confusion}

\begin{figure*}[!htbp]
    \centering
    \begin{subfigure}[b]{0.475\columnwidth}
        \centering
        \includegraphics[width=\columnwidth]{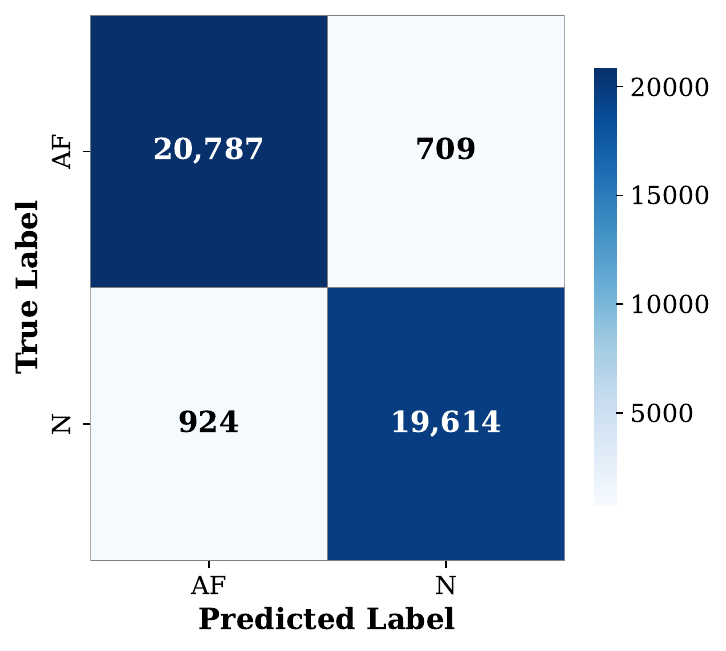}
        \caption{Raw counts}
        \label{fig:raw_counts}
    \end{subfigure}
    \begin{subfigure}[b]{0.475\columnwidth}
        \centering
        \includegraphics[width=\columnwidth]{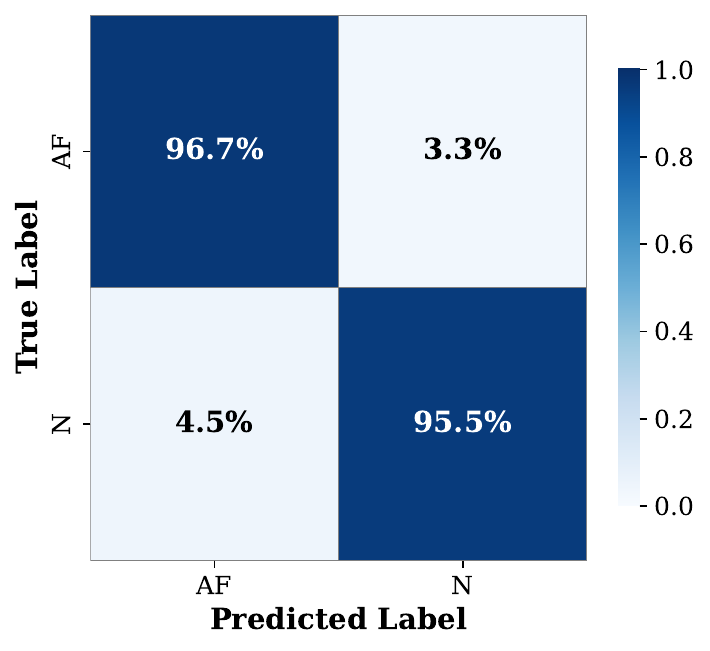}
        \caption{Row-normalized}
        \label{fig:row_normalized}
    \end{subfigure}
    
    \caption{\added{Confusion matrix for LTAFDB dataset. (a) Raw counts. (b) Row-normalized values showing recall per class. Out-of-distribution evaluation with 30 input windows.}}
    \label{fig:confusion_matrix}
\end{figure*}

\added{Figure~\ref{fig:confusion_matrix} presents a confusion matrix for the LTAFDB set that is out-of-distribution evaluated with 30 input windows. Since our approach uses a multi-label formulation with fractional targets, we approximate single-label assignments by taking the argmax of both ground-truth fractions (dominant rhythm) and predicted probabilities. The model achieves 96.7\% recall for AF and 95.5\% recall for N, with an overall accuracy of 96.1\%.}

\ifstandalone
  \bibliographystyle{elsarticle-num}
  \bibliography{bibfile}              
  \end{document}
\fi

\end{document}